\documentclass{article}

     \usepackage[preprint,nonatbib]{neurips_2020}

\usepackage[utf8]{inputenc} 
\usepackage[T1]{fontenc}    
\usepackage{hyperref}       
\usepackage{url}            
\usepackage{booktabs}       
\usepackage{amsfonts}       
\usepackage{nicefrac}       
\usepackage{microtype}      
\usepackage{xcolor}
\usepackage{graphicx}
\usepackage{amsmath}
\usepackage{amssymb}
\usepackage{subcaption}
\usepackage[ruled,vlined]{algorithm2e}

\title{Visual Transfer for Reinforcement Learning \\via Wasserstein Domain Confusion}

\author{
  Josh Roy\thanks{\texttt{joshnroy.github.io}} \\
  Department of Computer Science\\
  Brown University\\
  Providence, RI 02912 \\
  \texttt{joshnroy@gmail.com}\\
  \And
  George Konidaris \\
  Department of Computer Science\\
  Brown University\\
  Providence, RI 02912 \\
}

\begin{document}

\maketitle

\begin{abstract}
We introduce Wasserstein Adversarial Proximal Policy Optimization (WAPPO), a novel algorithm for visual transfer in Reinforcement Learning that explicitly learns to align the distributions of extracted features between a source and target task. WAPPO approximates and minimizes the Wasserstein-1 distance between the distributions of features from source and target domains via a novel Wasserstein Confusion objective. WAPPO outperforms the prior state-of-the-art in visual transfer and successfully transfers policies across Visual Cartpole and two instantiations of 16 OpenAI Procgen environments.
\end{abstract}

\section{Introduction}

Deep Reinforcement Learning (RL) has enabled agents to autonomously solve difficult, long horizon problems such as Atari games from pixel inputs~\cite{rlbook, atari}. In high risk domains such as robotics and autonomous vehicles, RL agents lack the ability to learn from their mistakes without destroying equipment or harming humans. Instead, it would be preferable to train in simulated domains and transfer to the real world. However, Deep model-free Reinforcement Learning agents trained on one environment often fail on environments that require solving the same underlying problem but different visual input~\cite{coinrun, procgen}. When transferring from a simulated source domain to a real-world target domain, this is known as the reality gap~\cite{sim2real1, dr1, cad2rl}. More generally, this challenge is known as generalization in Reinforcement Learning and is studied by transferring between simulated domains~\cite{coinrun, procgen, rlgeneralization1, rlgeneralization2, rlgeneralization3}.

The main reason for such difficulty in generalization is the tendency of deep networks to overfit to a single task~\cite{rlgeneralization1, rlgeneralization2, coinrun, humanpriors}. For example, an agent that learns to balance the cartpole depicted on the left of Figure \ref{fig:cartpole-samples} will fail to generalize to the cartpole on the right due to visual differences alone. In deep supervised learning, this can be addressed by smoothing the learned function using methods such as data augmentation~\cite{dataaugmentation}, dropout~\cite{dropout}, and regularization~\cite{regularization}. However, these methods are insufficient for generalization in Deep Reinforcement Learning~\cite{coinrun}. Unlike supervised learning, where all datapoints are drawn from the same underlying distribution, the problem of transferring between differing RL tasks is akin to that of supervised domain adaptation where datapoints from source and target tasks are drawn from different distributions~\cite{adda, dann, domainconfusion}.

Previous work such as Domain Randomization or Meta-Learning takes a brute force approach to this problem, training on many source domains before fine tuning on the target domain for $0$ or $k$ episodes, referred to as zero-shot or $k$-shot transfer, respectively. These methods require a large number of source domains before they can transfer or adapt to the target domain. Domain Randomization learns a representation sufficient to solve all source domains, implicitly aligning the distributions of features extracted from each domain. It then assumes the features extracted from the target domain will fall into a similar distribution~\cite{dr1, openaidexterity, openaicube}. Some recent work has attempted to more explicitly align the feature distributions across domains but assumes access to pairwise correlations of states across domains~\cite{robustdr, pairedreconstructionmapping, weakpairwisedomainadaptation}. Other work relies upon style-transfer networks to ``translate'' target domain states to source domain states before processing~\cite{vrgoggles}. This adds additional computational complexity during both training and inference and relies upon the quality of the translation network. Further work attempts to learn the causal structure underlying such a family of visual Markov Decision Processes (MDPs), leading to a causal representation independent of domain and successful transfer~\cite{causalblock}. Both translation and causal approaches require modeling of variables unrelated to the task given, leading to additional complexity~\cite{vrgoggles, causalblock}.

We introduce an algorithm that explicitly learns to align the distributions of extracted features between a source and target task without adding additional computation during inference or assuming access to pairwise correlations of observations. By simultaneously training an RL agent to solve a source task and minimize the distance between distributions of extracted features from the source and target domains, our algorithm enables seamless transfer to the target domain. Specifically, we train an adversary network to approximate the Wasserstein-1 distance (Earth Mover's Distance) between the distributions of features from source and target tasks, and an RL agent to minimize this distance via a novel Wasserstein Confusion objective while solving the source task~\cite{wgan}. The Wasserstein-1 distance between distributions can be intuitively described as the effort required to align two probability distributions by transporting their mass~\cite{emd}. As shown in \cite{wgan}, minimizing this distance allows adversarial alignment methods to succeed where other distance metrics such as Jensen-Shannon divergence fail. Our algorithm outperforms the prior state-of-the-art in visual transfer and successfully transfers policies across Visual Cartpole, a visual variant of the standard cartpole task where colors are changed across domains~\cite{gym}, and two varieties of 16 OpenAI Procgen environments~\cite{procgen}.

\section{Background and Related Work}

 Reinforcement Learning is concerned with sequential decision making~\cite{rlbook}: an RL agent exists within a world (environment) and must take an action $a$ based on some information about the world (state) $s$. This causes the environment to provide the agent with the next state $s'$ and a reward $r$. The agent's goal is to learn a policy $\pi$ mapping states to actions that maximizes the expected sum of rewards $\mathbb{E}[\sum_t \gamma^t r_t],$ where $\gamma \in [0, 1)$ is a discount factor that weights the importance of short and long term rewards. Return is defined as the sum of cumulative rewards. The interactions between the agent and the environment are modeled as a Markov Decision Process (MDP) defined as a 5-tuple$(S, A, T, R, \gamma)$ where $S$ is the set of states, $A$ is the set of actions, $T$ is a function mapping from state and action to next state, $R$ is a function mapping from state and action to reward, and $\gamma$ is the discount factor.

Model-free Deep RL uses neural networks to predict both the value (expected future reward) of a state and the optimal action to take. Proximal Policy Optimization (PPO) is a state-of-the-art model-free policy gradient algorithm for Deep RL~\cite{ppo}. It parameterizes a policy $\pi_\theta(a | s)$ and a value function $V_\theta(s)$ that share the majority of their weights, splitting after the ``feature extraction'' section of the network. The value network is trained to minimize the mean square error $\mathcal{L}_\text{value} = \frac{1}{n} \sum_{i=1}^n (V(s) - R)^2,$ where $R$ is the return. The policy network is trained to minimize $\mathcal{L}_\text{policy} = -\hat{\mathbb{E}}_t[\min(r_t(\theta) \hat{A}_t, \text{clip}(r_t(\theta), 1 - \epsilon, 1+\epsilon)\hat{A}_t],$ where $\hat{\mathbb{E}}$ is the empirical expected value at timestep $t$, $\hat{A}_t$ is the empirical advantage at timestep $t$, $r(\theta) = \frac{\pi_\theta(a_t | s_t)}{\pi_{\theta_\text{old}}(a_t | s_t)}$ is the ratio of taking action $a_t$ given state $s_t$ between the current and previous policies, and $\epsilon$ is a small hyperparameter. The two function approximators are trained jointly, and their combined loss is $\mathcal{L}_\text{PPO} = \mathcal{L}_\text{policy} + \mathcal{L}_\text{value}$.

\subsection{Transfer in Reinforcement Learning}

Transfer in Reinforcement Learning has been a topic of interest far before the recent advent of Deep Neural Networks and Deep Reinforcement Learning. Work on transfer is separated into questions of dynamics, representation, and goals with environments differing in their states, actions, transition function, or reward function, respectively~\cite{taylor2009transfer, lazaric2012survey}. A policy that transfers perfectly is one that trains on a source MDP and achieves target reward equivalent to that of an agent trained on a target MDP.

The most popular approach in $k$-shot transfer for RL is Meta Reinforcement Learning. These methods optimizes a Deep RL agent's parameters such that it can rapidly learn any specific task selected from a set of tasks. Specifically, the agent first trains on $n$ source tasks and then trains on the $n+1$th task for $k$ episodes before measuring performance on the $n+1$th task~\cite{maml, reptile}. While Meta RL is an interesting and relevant field of study, it requires the ability to fine tune by training in target domains.

Domain Randomization is the most popular approach to zero-shot transfer in RL. In Domain Randomization, an RL agent trains on a set of $n$ source tasks, implicitly learning a representation and policy sufficient to transfer to zero-shot transfer to the $n+1$th task~\cite{dr1, openaidexterity, openaicube}. For successful transfer, all $n+1$ tasks must be drawn from the same distribution. Furthermore to enable this implicit learning, $n$ is required to be sufficiently large that the agent is unable to memorize domain-specific policies.

Some work directly addresses dynamics transfer~\cite{hip1, hip2, hip3} by parameterizing the transition function of an MDP and learning a conditional policy that can transfer between such tasks. Other work generates a curriculum to learn generalizable policies that can adapt to MDPs with differing dynamics~\cite{rewardguided}. This work is complementary to WAPPO which focuses on appearance-based transfer.

\subsection{Visual Transfer in Reinforcement Learning}

Visual transfer takes place within a family $\mathcal{M}$ of related Block MDPs $M \in \mathcal{M}$ each defined by a $6$-tuple $(S, \mathcal{A}, \mathcal{X}, p, q, R)$ where $S$ is an unobserved state space, $\mathcal{A}$ is an action space, $\mathcal{X}$ is an observation space, $p(s' | s, a)$ is a transition distribution over the next state $s'$ based on the previous state $s$ and action $a$, $q(x | s)$ is an emission function that represents the probability of an observation $x$ based on a hidden state $s$, and $R(s, a)$ is a reward function that maps from a state and action to a reward~\cite{blockmdp, causalblock}. The emission function $q$ and the observation space $\mathcal{X}$ are the only quantities that change across Block MDPs within a family. Block MDPs are similar to POMDPs but have emission functions rather than observations functions~\cite{blockmdp, oldpomdp}. Both functions both map from hidden states to observations but emission functions generate observations that are definitionally Markov~\cite{blockmdp}.

To transfer across different visual domains, Robust Domain Randomization (RDR)~\cite{robustdr} aims to learn a domain-agnostic representation by training an agent to solve $n$ source domains while minimizing the Euclidian distance between internal representations across domains. It then attempts to zero-shot transfer the $n+1$th domain. This method shows success in tasks where the background color is the only varied property, but the visual randomization of complex tasks such as OpenAI Procgen~\cite{procgen} is far higher. When the training data contains pairs of samples with the same semantic content in each domain, minimizing the Euclidian distance will align the distributions of representations for each domain. Without paired data, this will incorrectly align samples with different hidden states, leading to unreliable representations. Similar works from Gupta et al. and Tzeng et al. learn a pairing between source and domain images and minimize paired distance, aligning the feature distributions across domains~\cite{pairedreconstructionmapping, weakpairwisedomainadaptation}. However, these methods assume a pairing of images across domains exists within the data given to the agent, which is unlikely to occur by chance in complex visual domains.

Other work uses causal inference to learn Model-Irrelevance State Abstractions (MISA)~\cite{causalblock} for transfer between Block MDPs. MISA successfully transfers between RL tasks with low-dimensional states and visual imitation learning tasks with changing background color~\cite{causalblock} but not visual RL tasks. Since this algorithm relies upon reconstructing observations, it must model factors that may not be relevant to the task. In visual tasks, MISA minimizes mean squared error between observations and reconstructions, which is demonstrated to ignore small objects due to their low effect on the error signal~\cite{disappearingreconstruction}. In visual RL, small objects such as the player character are essential to solving the task.

Work in supervised domain adaptation~\cite{cycada} and style transfer~\cite{cyclegan, styletransfersurvey, styletransfer1} translates images to different ``styles'' such as those of famous painters by training Generative Adversarial Networks (GANs) to map from an input image to a semantically equivalent but stylistically different output image. VR Goggles for Robots~\cite{vrgoggles} aims to transfer RL policies across visually differing domains by using a style transfer network to translate target domain images into the source domain. This enables an RL agent trained in the source domain to act in the target domain but adds additional computational and algorithmic complexity during inference time and relies heavily upon the success of the style-transfer network~\cite{vrgoggles} as the policy does not train on translated images. As the style transfer network minimizes the mean absolute error and mean squared error between real and generated images, it prioritizes modeling factors that utilize a higher number of pixels. In complex visual RL tasks, important factors, such as the player character, occupy a small number of pixels and recieve low priority~\cite{disappearingreconstruction}. While style-transfer adds computational complexity and relies upon image reconstruction, other work in GANs and supervised domain transfer aligns distributions without these drawbacks.

\subsection{Adversarial Distribution Alignment}

Prior work in GANs~\cite{gan} shows that adversarial methods can align arbitrary distributions of data. Specifically, given samples drawn from a ``real'' distribution $P_r$ a GAN learns a mapping from noise sampled from a Gaussian distribution $P_z$ to samples from a ``fake'' distribution $P_f$. It jointly trains an adversary network to classify samples as real or fake and a generator network to ``fool'' the adversary, minimizing the Jensen-Shannon divergence (JS divergence) between the distributions~\cite{gan}.

Some domain adaptation methods in supervised learning uses a GAN-like adversary to align a classifier's internal representations across domains and solve the supervised classification analogue of visual transfer in RL~\cite{adda, dann, domainconfusion}. They each introduce and minimize different distribution alignment objectives based on minimizing the classification accuracy of the adversary network~\cite{domainconfusion, adda}.

GAN-like adversarial algorithms align distributions but are unstable and prone to failures such as mode collapse~\cite{wgan}. Furthermore, minimizing the JS divergence between two distributions has been shown to fail is certain cases, such as aligning two uniform distributions on vertical lines~\cite{wgan}. Wasserstein GANs solve both of these problems by changing the adversarial classifier into an adversarial critic $f$ that estimates the Wasserstein-1 distance $W(P_r, P_f)$ between real and fake distributions $P_r$ and $P_f$~\cite{wgan}. Minimizing Wasserstein-1 distance is more stable and empirically shown to avoid mode collapse~\cite{wgan}. Though directly estimating the Wasserstein-1 distance is infeasible~\cite{wgan}, the Kantorovich-Rubinstein duality~\cite{wassersteinduality} provides a reparameterization that is directly computable. The gradient of the generator $G$ with weights $\theta$ is defined as $\nabla_\theta W(P_r, P_f) = - E_{z \sim P_z}[\nabla_\theta f(G(z))]$.

\section{Wasserstein Adversarial Proximal Policy Optimization}

\begin{figure}[h]
\centering
\includegraphics[width=0.2\textwidth]{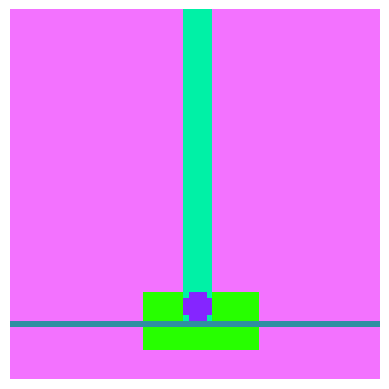}
\includegraphics[width=0.2\textwidth]{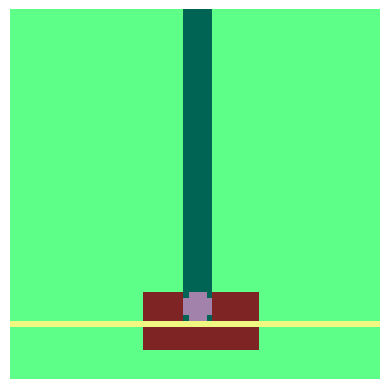}
\caption{Example Visual Cartpole Source Domain (Left) and Target Domain (Right). The two domains contain the same hidden states, actions, rewards, and transitions but differ in their emission functions resulting in differing observations.}
\label{fig:cartpole-samples}
\end{figure}

Our novel algorithm Wasserstein Adversarial Proximal Policy Optimization (WAPPO) transfers from any source task $M_s$ to any target task $M_t$ where both $M_t, M_s$ are Block MDPs drawn from a family $\mathcal{M}$. For any two such tasks, the observation spaces $\mathcal{X}_s, \mathcal{X}_t$ and emission functions $q_s, q_t$ are different, but the hidden state space $S$, action space $\mathcal{A}$, transition function $p$, and reward function $R$ are identical. The Visual Cartpole environments shown in Figure \ref{fig:cartpole-samples} illustrate difference in emission function; the source domain has a green cart and pole as on a pink background and blue track while the target domain has a brown cart with a green pole on a green background and yellow track. However, both environments are defined by the same hidden states (position of the cart and pole), actions (push the cart left or right), rewards ($+1$ for keeping the pole upright at each timestep), and transitions (simulation of physics). Furthermore, the optimal policy that balances the pole is solely a function of the hidden states representing the position of the cart and pole and not of the colors of the objects.

In model-free Deep RL, the agent does not learn the state of its MDP. Instead, by learning a policy that maximizes its reward, the agent implicitly learns a representation function $h_\theta$ parameterized by the first few layers of the RL network that maps from observations $x \in \mathcal{X}$ to internal representations $\mathfrak{r} \in \mathfrak{R}$. In the Block MDPs defined above, an optimal policy depends solely on the hidden state space $S$. Thus, an optimal representation function $h^*$ will map observations $x \in \mathcal{X}$ to internal representations $\mathfrak{r} \in \mathfrak{R}$ that contain no more information than their corresponding hidden states.

Similarly, there exist optimal representation functions $h_s^*, h_t^*$ for source and target tasks $M_s, M_t \in \mathcal{M}$. Since the hidden state spaces of $M_s, M_t$ are identical, the optimal policies and representation spaces $\mathfrak{R}_s, \mathfrak{R}_t$ are identical. Thus, there exists a representation function $h_{(s, t)}^*$ that maps from both $\mathcal{X}_s$ and $\mathcal{X}_m$ to a representation space $\mathfrak{R} = \mathfrak{R}_s = \mathfrak{R}_t$ that is sufficient to solve both source and target tasks. 

The goal of Wasserstein Adversarial PPO is to learn a representation function $h_{\theta, (s, t)}$ that approximates the optimal representation function $h_{(s, t)}^*$.  It is given the ability to train within $M_s$ and access a buffer of observations sampled from $M_t$. Note that the observations from $M_t$ are not paired with observations from $M_s$ and are not sequential. Further, WAPPO does not have access to rewards from $M_t$. Given that Block MDPs $M_s, M_t$ both belong to the family $\mathcal{M}$, they share hidden states, transition function, and reward function but differ in observations. Therefore, an agent with a domain-agnostic representation will be able to learn a policy in $M_s$ and transition seamlessly to $M_t$. We define the representation function $h_\theta$ as the first few layers of the RL network with parameters $\theta$. To learn a domain-agnostic representation, Wasserstein Adversarial PPO jointly learns to solve $M_s$ while using an adversarial approach to align the distributions of representations from $M_s$ and $M_t$.

Specifically, Wasserstein Adversarial PPO trains on a dataset $D$ defined as $\{(x_s, x_t, a_s, r_s)_1, ... (x_s, x_t, a_s, r_s)_m\}$ where $m$ is the length of the dataset, $x_s \in \mathcal{X}_s$ is an observation from the source MDP, $x_t \in \mathcal{X}_t$ is an observation from the target MDP, $a_s$ is an action from the source MDP, and $r_s$ is a reward from the source MDP. Note that $x_s, a_s, r_s$ correspond to the same timestep from the source MDP: the agent takes action $a_s$ from state $x_s$, receives reward $r_t$, and transitions to the next state. $x_t$ does not correspond to the same timestep as the other variables. Instead, it is drawn randomly from a buffer of observations from the target MDP. WAPPO uses the output of an intermediate layer, denoted by $h_\theta(x)$ as a latent representation $\mathfrak{r}$ of the observation $x$. To enforce that this latent representation $x$ is agnostic to domain, WAPPO approximates $h^*$ by minimizing the Wasserstein-1 distance between the distributions of $h_\theta(x_s)$ and $h_\theta(x_t)$.

Similar to supervised adversarial domain adaptation algorithms~\cite{adda, dann, domainconfusion}, WAPPO consists of two networks: the RL network and the adversary network shown in Figure \ref{fig:network-arch} in the Appendix. The RL network learns a latent representation which is used to compute the best next action and the value of each observation. This latent representation should be identical across source and target domain. The adversary network takes the latent representation as input and is trained to distinguish between source and target tasks. The policy network both maximizes performance on the source task and minimizes the adversary's ability to identify the domain. Specifically, the RL network minimizes $$\mathcal{L}_\text{WAPPO} = \mathcal{L}_\text{PPO} + \lambda \mathcal{L}_\text{Conf},$$ where $\mathcal{L}_\text{PPO}$ is the PPO loss and $\mathcal{L}_\text{Conf}$ is the loss term that maximally confuses the critic and aligns the distributions of source and target observations.

The Wasserstein GAN~\cite{wgan} approximates the Wasserstein distance between real and fake distributions using a neural network. Additionally, it defines a derivative for this distance that is used in optimizing the loss term for the generator, demonstrating higher stability than the standard GAN loss.

While the Wasserstein GAN loss term seems to align exactly with $\mathcal{L}_\text{Conf}$, it has one key difference. It assumes that one distribution is fixed, which is not true of domain adaptation. The goal of domain adaptation is to align two distributions both parameterized by $\theta$. Specifically, we wish to align the distribution of extracted features from the source domain $P_{s}^{h_\theta}$ with the extracted features of the target domain $P_{t}^{h_\theta}$. Note that it is not possible to directly sample from $P_s^{h_\theta}, P_t^{h_\theta}$. Instead, we sample from these distributions by first sampling from the distribution of observations, $P_s, P_t$ and then mapping to the representation by applying $h_\theta$. Thus, the Wasserstein distance is defined as 
\begin{equation}
W(P_{s}, P_{t}) = E_{x \sim P_{s}}[f(h_\theta(x))] - E_{x \sim P_{t}}[f(h_\theta(x))],
\end{equation}
where $f$ is the adversarial critic and the gradient is defined as:
\begin{align*}
    \nabla_\theta W(P_{s}, P_{f_{t}}) & = \nabla_\theta [E_{x \sim P_{s}}[f(h_\theta(x))] - E_{x \sim P_{t}}[f(h_\theta(x))]]\\
    & = \nabla_\theta E_{x \sim P_{s}}[f(h_\theta(x))] - \nabla_\theta E_{x \sim P_{t}}[f(h_\theta(x))]
\end{align*}
\begin{equation}
\nabla_\theta W(P_{s}, P_{t}) = E_{x \sim P_{s}}[\nabla_\theta f(h_\theta(x))] - E_{x \sim P_{t}}[\nabla_\theta f(h_\theta(x))].
\end{equation}

Moving the gradient inside the expectation is shown to be correct in the proof of Theorem 3 of \cite{wgan}.

\section{Experimental Results}

We validate our novel Wasserstein Confusion loss term and WAPPO algorithm on 17 environments: Visual Cartpole and both the easy and hard versions of 16 OpenAI Procgen environments. To evaluate the ability of the Wasserstein Confusion loss term to align distributions of features across environment and enable successful transfer, we examine how an RL agent trained using WAPPO on a source domain performs on a target domain. For each environment evaluated, the agent trains using WAPPO with full access to the source domain and a buffer of observations from the target domain. The agent does not have access to rewards from the target domain. We compare WAPPO's transfer performance with that of three baselines: PPO, PPO with the feature matching loss as described by Robust Domain Randomization, the prior state of the art for feature alignment in RL~\cite{robustdr}, and PPO with VR Googles, the prior state of the art for domain adaptation in RL~\cite{vrgoggles}. 

Robust Domain Randomization originally trains on $n$ source domains while matching their features before attempting to transfer zero-shot to the target domain. It's main contribution is a feature alignment loss term. By minimizing this term and aligning the distributions of features extracted from the $n$ source domains, it hopes that the distribution of features from the target domain will also be aligned. We directly evaluate RDR's ability to match distributions of features by training an RL agent on one source domain and evaluating on one target domain while minimizing it's feature alignment loss using observations from each domain. As in the zero-shot setting, the agent's performance on the target domain is proportional to the alignment of features from source and target domains. Furthermore, this enables a direct comparison between the feature alignment loss used in Domain Randomization and our feature alignment loss, Wasserstein Confusion.

VR Goggles trains an RL agent on the source domain and a style-transfer network between target and source domain~\cite{vrgoggles}. During evaluation on the target domain, it translates images to the style of the source domain before applying the RL agent's trained policy. As VR Goggles utilizes a pre-trained source agent rather than a new RL algorithm, we report the target performance on Figures \ref{fig:cartpole-training}, \ref{fig:procgen-training-easy}, and \ref{fig:procgen-training-hard} as a horizontal line. We use the baseline PPO agent as the pre-trained source agent.

Each experiment's performance is reported across multiple trials with identical random seeds across algorithms. Visual Cartpole and Procgen Easy are evaluated across 5 trials and Procgen Hard is evaluated across 3 trials. There is one source domain and one target domain per trial.

\subsection{Visual Cartpole}

We first demonstrate performance on a simple environment, Visual Cartpole, a variant on the standard Cartpole environment~\cite{gym} where observations are RGB images of the Cartpole rather than position and velocity. Color of the cart, pole, background, track, and axle are varied across domains, converting the original MDP into a family of Block MDPs where different emission functions correspond to different colors. A sample source and target environment are shown in Figure \ref{fig:cartpole-samples}.

\begin{figure}[h]
\centering
\begin{subfigure}{0.5\textwidth}
\includegraphics[width=0.9\linewidth]{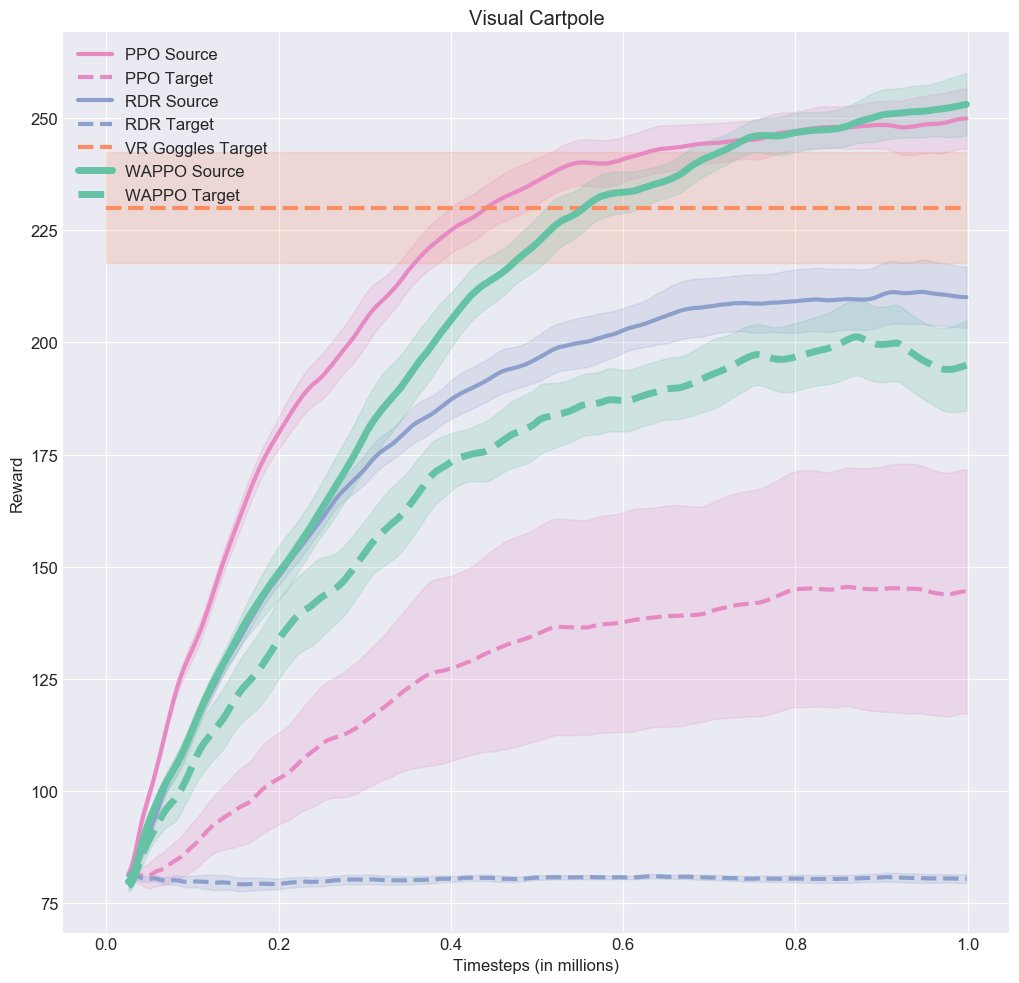}
\caption{Training Graph.}
\label{fig:cartpole-training}
\end{subfigure}
\begin{subfigure}{0.3\textwidth}
\begin{center}
\includegraphics[width=0.45\linewidth]{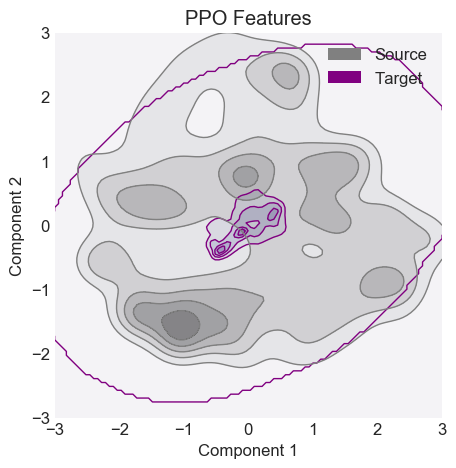}
\includegraphics[width=0.45\linewidth]{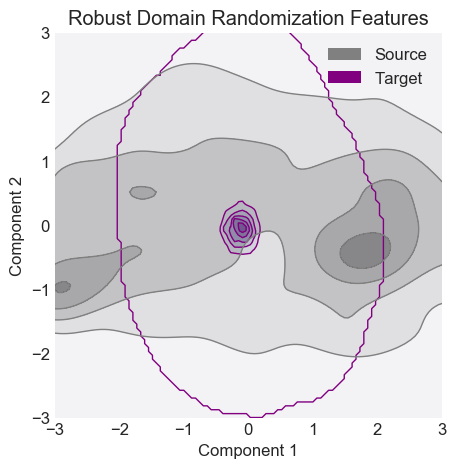}\\
\includegraphics[width=0.9\linewidth]{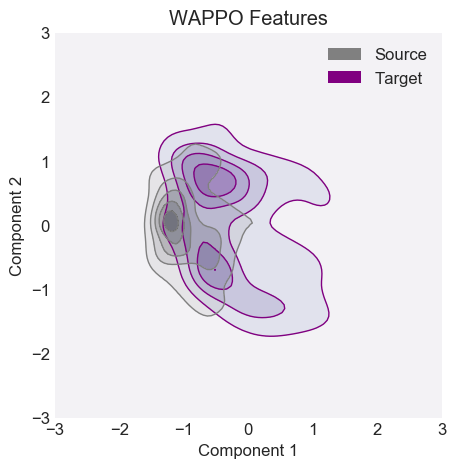}
\end{center}
\caption{Feature Distributions.}
\label{fig:cartpole-distributions}
\end{subfigure}
\caption{Visual Cartpole Training Graph (a) and Feature Distributions (b): Solid and dashed lines indicate source and target reward, respectively. Green indicates WAPPO, pink indicates PPO, blue indicates RDR, and orange indicates VR Goggles. Shading shows standard deviation across 5 trials. Gray and purple indicate source and target features, respectively. WAPPO has both higher target reward and better alignment of source and target features than other methods.}
\end{figure}

As shown in Figure \ref{fig:cartpole-training}, Wasserstein Adversarial PPO far outperforms both PPO and PPO using RDR's feature matching loss. As shown in Figure \ref{fig:cartpole-distributions}, the target features extracted by both PPO and RDR lie within a very small subset of their corresponding source features. This dense clustering implies that the target features are not expressive compared to the source features, leading to low transfer performance. When using Wasserstein Adversarial PPO, the target features are clustered at similar density with the source features, demonstrating their distributional alignment and leading to higher reward. The density plots in Figure \ref{fig:cartpole-distributions} are generated by reducing the features space to 2 dimensions via Principle Component Analysis (PCA)~\cite{pca} and estimating a density via Gaussian kernel density estimation~\cite{gaussiandensityestimation}. While VR Goggles outperforms WAPPO and the other algorithms in Visual Cartpole, it does not in complex visual environments such as Procgen~\cite{procgen}. VR Goggles is able to perform well in Visual Cartpole as the background color is the only visual distraction and the cartpole system occupies a majority of the visual observation. We do not visualize distributions of features for VR Goggles as it operates on the observation space rather than the image space.

\subsection{OpenAI Procgen}

The remaining 16 environments are from OpenAI's Procgen Benchmark~\cite{procgen}. This benchmark originally provides agents $n$ source domains to train on and tests their generalization to a different target domain. The domains differ according to their emission functions, dynamics, and states. As this work focuses on visual domain adaptation in Reinforcement Learning, we modify the environments such that each environment is a family of Block MDPs. All domains in a particular environment have identical state spaces, transition functions, and reward functions but unique emission functions. This decouples the different types of transfer and solely focuses on evaluating visual transfer. As in Visual Cartpole, we train the agent in one source domain with access to observations from the target domain and test the agent's performance on the target domain. The environments vary in the difficulty of both the source task and generalizing to a new domain. Chaser, Heist, Maze, and Miner revolve around navigating in $2$-dimensional mazes of varying size. This is a difficult task for model-free deep RL agents as they cannot plan a path betwen maze locations and causes difficulty for all of the agents evaluated~\cite{rlmazes}. Furthermore, there are two different instantiations of each environment: an easy version and a hard version. The easy version of each game has less variation across all factors, such as colors, layouts, and goal locations~\cite{procgen}. In certain games, the easy version provides additional hints to the agent that simplify the task, such as an arrow pointing in the direction of the goal.

We report both the reward on each task and the normalized return across all tasks. Normalized return is calculated by averaging the normalized return on each task $R_\text{norm} = (R - R_\text{min}) / (R_\text{max} - R_\text{min})$, where $R$ is the vector of rewards and $R_\text{min}, R_\text{max}$ are the minimum and maximum returns for each environment as defined by ~\cite{procgen}. Note that the minimum and maximum return for each environment are defined by their theoretical limits rather than the minimum and maximum performance of any algorithm. This quantity measures the overall source and target performance for each algorithm. 

\begin{figure}[h]
\centering
\includegraphics[width=\textwidth]{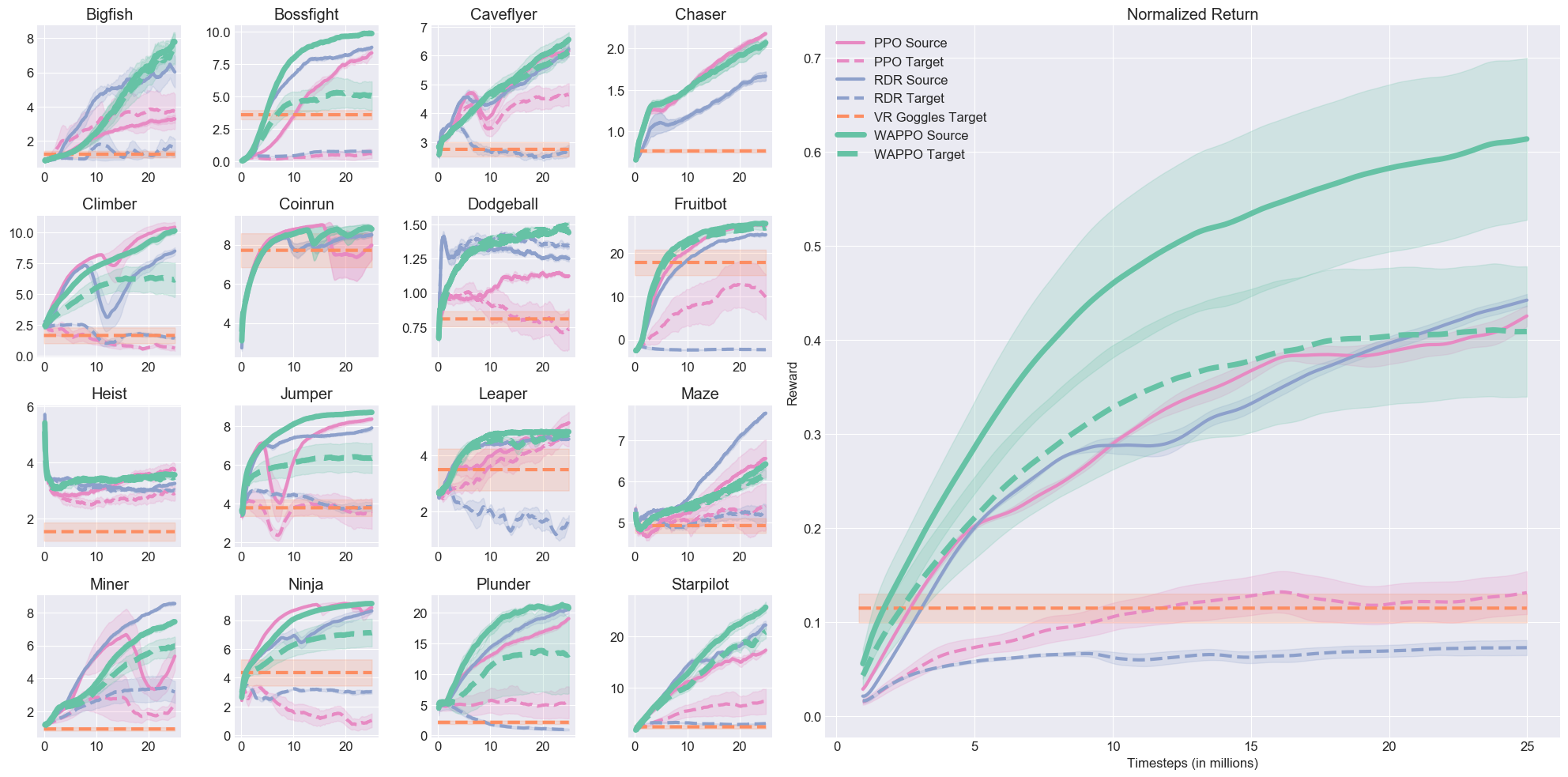}
\caption{Procgen (Easy) Training Graph. Solid lines and dashed lines indicate source and target reward, respectively. Green indicates WAPPO, blue indicates PPO, pink indicates RDR, and orange indicates VR Goggles. WAPPO matches or outperforms the other algorithms across all environments.}
\label{fig:procgen-training-easy}
\end{figure}

As shown in Figure \ref{fig:procgen-training-easy}, WAPPO's target performance exceeds that of the other three algorithms in $14$ of the $16$ easy environments. In Coinrun and Chaser, the target performance of all algorithms approximately matches their training performance, demonstrating the environments' lack of transfer difficulty. In Chaser, WAPPO's performance exceeds that of Robust Domain Randomization and matches that of PPO. In Coinrun, WAPPO matches the performance of RDR and exceeds PPO.

\begin{figure}[h]
\centering
\includegraphics[width=\textwidth]{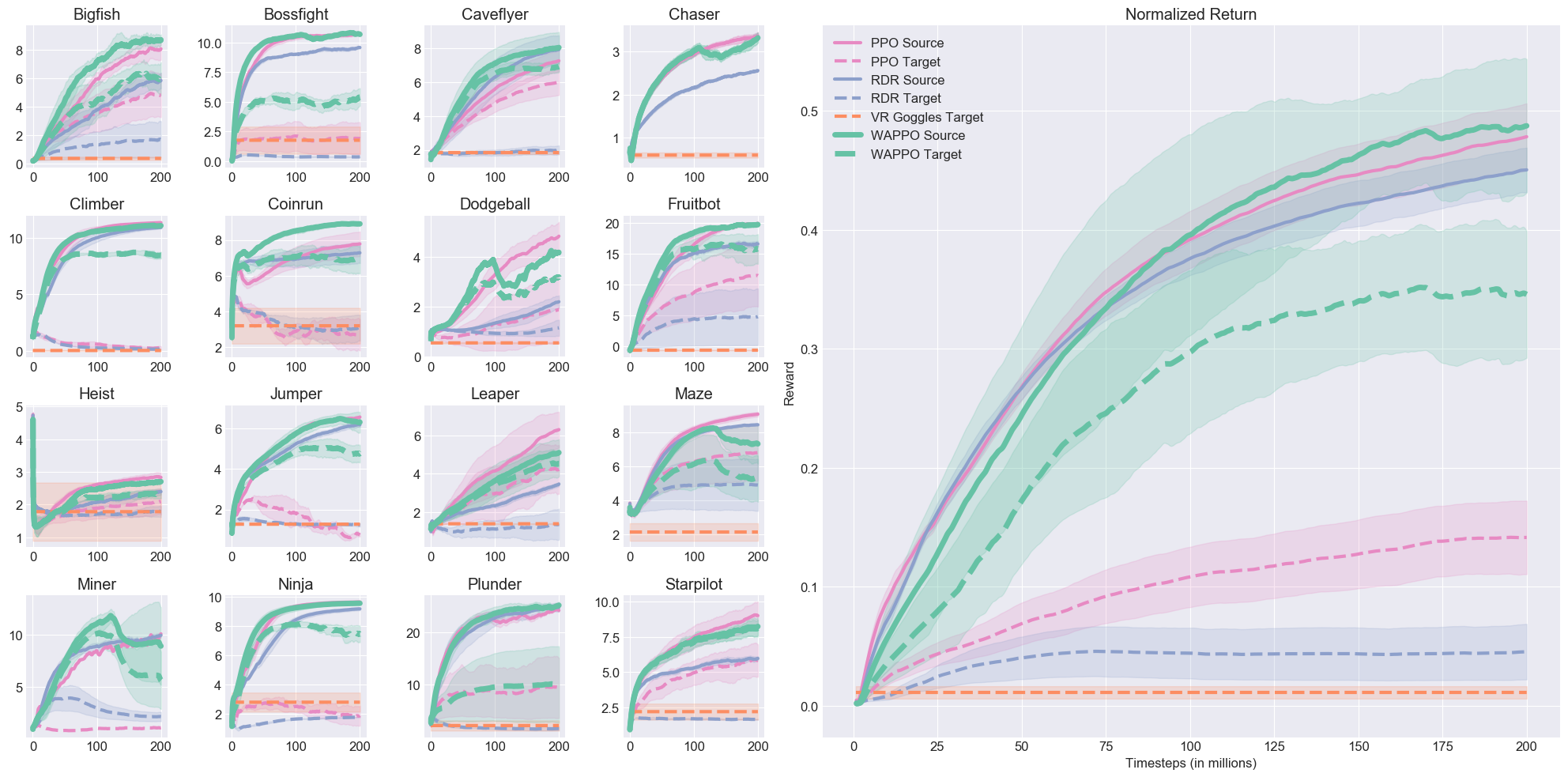}
\caption{Procgen (Hard) Training Graph. Solid lines and dashed lines indicate source and target reward, respectively. Green indicates WAPPO, blue indicates PPO, pink indicates RDR, and orange indicates VR Goggles. WAPPO matches or outperforms the other algorithms across all environments except Maze which primarily tests path planning ability rather than visual transfer.}
\label{fig:procgen-training-hard}
\end{figure}

These results are mirrored in the hard versions of the Procgen environments, as shown in Figure \ref{fig:procgen-training-hard}. WAPPO's target performance exceeds that of the other algorithms in $12$ of the $16$ hard environments. On Plunder, Chaser, and Leaper, WAPPO matches the performance of PPO and outperforms other algorithms. Chaser has minimal visual variation, allowing all algorithms match their source and target performance. On Maze, WAPPO performs worse than PPO but better than other visual transfer algorithms. Maze tests an agent's ability to solve mazes and has few visual differences across domains. Maze solving is difficult for model-free Deep RL agents due to their lack of ability to plan a path between locations~\cite{rlmazes}. As shown in Figures \ref{fig:procgen-training-easy} and \ref{fig:procgen-training-hard}, WAPPO's normalized return on easy and hard Procgen environments far exceeds that of prior algorithms, demonstrating its superior ability to generalize to a target domain. Furthermore, aligning source and target distributions allows WAPPO to ignore distracting visual details in the source domain and achieve higher source domain reward.

\section{Conclusion}

We address the difficulty in transferring a learned policy across differing visual environments by explicitly learning a representation sufficient to solve both source and target domains. Previous approaches have added additional inference-time complexity~\cite{vrgoggles}, relied upon pairwise observations~\cite{pairedreconstructionmapping, weakpairwisedomainadaptation}, or ignored fine details by relying upon image reconstruction losses~\cite{vrgoggles, causalblock}. We have introduced a new method, WAPPO, that does not. Instead, it uses a novel Wasserstein Confusion objective term to force the RL agent to learn a mapping from visually distinct domains to domain-agnostic representations, enabling state-of-the-art domain adaptation performance in Reinforcement Learning. We validate WAPPO on 17 visual transfer environments where our agent both achieves higher reward in the target MDP and better matches distributions of representations across domains.

\pagebreak
\section*{Broader Impact}

Wasserstein Adversarial PPO will enable RL agents to better train in simulation and successfully be applied to real-world scenarios. This transfer will be especially useful in fields such as robotics and autonomous vehicles. Thus, it inherits the broader impacts of Reinforcement Learning as a whole and those of work that makes RL more practical. By increasing the generalization of trained RL policies, Wasserstein Adversarial PPO works toward enabling robots to help humans in daily life, ranging from aiding elders in their homes to making autonomous deliveries. This also furthers the applicability of RL in fields such as factory/warehouse automation and disaster robotics by enabling RL agents to transfer across different environments. Although Wasserstein Adversarial PPO enables Deep RL to be applied in a wider variety of domains, it does not solve safety guarantees inherent to the use of Neural Networks. Specifically, there is no guarantee that a Deep RL agent (trained with or without the Wasserstein Confusion objective) will act correctly in all situations. As this may lead to undesirable behavior, we expect WAPPO to be utilized in systems that contain other safety mechanisms. These mechanisms include but are not restricted to power and force limiting, automatic breaking, and compliant robotic arms~\cite{collaborativerobotsafety, autonomousvehiclesafety}. WAPPO's contributions are complimentary to that of current and future safety mechanisms. When both used in the same system, safety mechanisms will detect and prevent failure cases and WAPPO will enable successful transfer to real-world scenarios.



\bibliography{bibliography}
\bibliographystyle{unsrt}

\newpage

\section*{Appendix}

\subsection*{Algorithm}

The goal of the RL network is to simultaneously minimize the PPO loss $\mathcal{L}_\text{RL}$ and the confusion loss $\mathcal{L}_\text{Conf}$. It samples observations, actions, and rewards from the source environment and observations from the target environment. If then computes and minimizes these losses. $f$ is a function that approximates the Wasserstein distance between the source observation and target observation distributions. Thus, it should be trained to convergence for every update of the RL network. As in Wasserstein GAN, it is optimized for $n_\text{critic}$ steps for each update of the RL network. This process is outlined in Algorithm \ref{alg:adversarial-ppo}. Note that we use the weight clipping method defined in \cite{wgan} rather than the gradient penalty method defined in \cite{wgangp} to directly test the effect of the novel Wasserstein Confusion loss term. We believe that combining Wasserstein Confusion with gradient penalty is a promising direction for future work.

\begin{algorithm}[h]
\SetAlgoLined
 \For{$t=0, ..., n_\text{timesteps}$}{
     \For{$j=0, ..., n_\text{critic}$}{
        Sample $\{\mathfrak{s}_{s, i}\}_{i=1}^m \sim P_{s}$ a batch from the source domain\\
        Sample $\{\mathfrak{s}_{t, i}\}_{i=1}^m \sim P_{t}$ a batch from the target domain buffer\\
        $\mathbb{g}_w \leftarrow \nabla_w [\frac{1}{m} \sum_{i=1}^m f_w(h_\theta (\mathfrak{s}_{s, i})) - \frac{1}{m} \sum_{i=1} f_w(h_\theta (\mathfrak{s}_{t, i}))]$\\
        $w \leftarrow w + \alpha \times $ RMSProp$(w, \mathbb{g}_w)$\\
     }
    Sample $\{\mathfrak{s}_{s, i}, a_{s, i}, r_{s, i}\}_{i=1}^m \sim P_{s}$ a batch from the source domain\\
    Sample $\{\mathfrak{s}_{t, i}\}_{i=1}^m \sim P_{t}$ a batch from the target domain buffer\\
    $\mathbb{g}_\theta \leftarrow \nabla_\theta [-\frac{1}{m} \sum_{i=1}^m f_w(h_\theta (\mathfrak{s}_{s, i})) + \frac{1}{m} \sum_{i=1} f_w(h_\theta (\mathfrak{s}_{t, i})) + \mathcal{L}_\text{RL}(\mathfrak{s}_{s, 1}, a_{s, 1}, r_{s, 1}, ..., \mathfrak{s}_{s, m}, a_{s, m}, r_{s, m})]$\\
    $\theta \leftarrow \theta - \alpha \times $ RMSProp$(\theta, \mathbb{g}_\theta)$
 }
 \caption{Wasserstein Adversarial PPO}
 \label{alg:adversarial-ppo}
\end{algorithm}

\subsection*{Implementation Details}

\begin{figure}[h]
    \centering
    \includegraphics[width=0.75\textwidth]{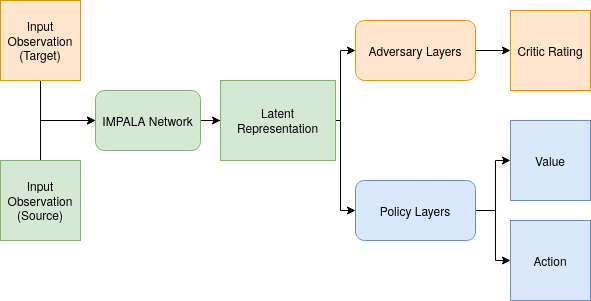}
    \caption{Network architecture.  Layers are represented as rounded rectangles. Blue indicates use in training the RL policy, orange indicates use in training the critic, and green indicates use in training both. Note that the network architecture mirrors that of domain confusion and randomization but is modified to work with Reinforcement Learning rather than Supervised Learning~\cite{domainconfusion,adda,dann}. The combination of the green and blue networks is identical in architecture to the IMPALA network used to benchmark the Procgen Environments and mirrors that used in Robust Domain Randomization~\cite{procgen,robustdr}. Only the green and blue networks were used when measuring the performance of PPO and Robust Domain Randomization.}
    \label{fig:network-arch}
\end{figure}

There are four algorithms that we implement: PPO, Robust Domain Randomization, VR Goggles, and WAPPO. All four are built off of PPO~\cite{ppo}. We use the high quality, open source implementation of PPO provided by OpenAI Baselines~\cite{baselines}. Furthermore, we use the neural architecture and learning rate provided by \cite{procgen} as a baseline. This architecture consists of the CNN component from the IMPALA network~\cite{impala}, which is shared between the policy and value prediction components of PPO. The value and policy networks then branch and each have one fully connected layer which outputs the policy or value, respectively. As in \cite{procgen}, we use a learning rate of $5 \times 10^{-4}$. We evaluated with both the Adam Optimizer~\cite{adam} and the RMSProp Optimizer~\cite{rmsprop} but find negligible difference.

For PPO, Robust Domain Randomization, and VR Goggles, the network does not have an adversary. As such, we used the green and blue sections depicted in Figure \ref{fig:network-arch} to train the PPO agent used in both these algorithms. The VR Goggles training process is the same as that used in \cite{vrgoggles}. Specifically, we collect a dataset of $2000$ source domain images and $2000$ target domain images and train the VR Goggles translation network with a learning rate of $2 \times 10^{-4}$ for $50$ epochs. As in \cite{vrgoggles}, we found no performance gain when using a larger dataset or training for more epochs. As \cite{vrgoggles} does not provide an implementation, we re-implement this method by adding their novel shift loss to the open source CycleGAN implementation~\cite{cyclegan}. For Robust Domain Randomization, we implement the regularization loss and use a regularization weight of $10$ and as described in \cite{robustdr}.

For WAPPO, we use the entire network depicted in Figure \ref{fig:network-arch}. The green and blue sections are optimized according to the PPO loss $\mathcal{L}_\text{PPO}$ and the green and orange sections are optimized according to the Wasserstein Confusion loss $\mathcal{L}_\text{Conf}$. Similarly to \cite{wgan}, we take $5$ gradient steps of the adversary network per step of the RL network. The adversarial critic network is made of $8$ dense layers of width $512$, separated by Leaky ReLU~\cite{leakyrelu} activation function.

\subsection*{Observations Across Visual Domains}

\begin{figure}[h]
\centering
\begin{subfigure}{0.35\textwidth}
\begin{center}
\includegraphics[width=0.4\textwidth]{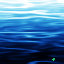}
\includegraphics[width=0.4\textwidth]{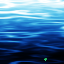}
\end{center}
\caption{Bigfish.}
\end{subfigure}
\begin{subfigure}{0.35\textwidth}
\begin{center}
\includegraphics[width=0.4\textwidth]{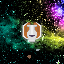}
\includegraphics[width=0.4\textwidth]{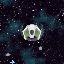}
\end{center}
\caption{Bossfight.}
\end{subfigure}\\
\begin{subfigure}{0.35\textwidth}
\begin{center}
\includegraphics[width=0.4\textwidth]{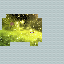}
\includegraphics[width=0.4\textwidth]{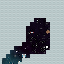}
\end{center}
\caption{Caveflyer.}
\end{subfigure}
\begin{subfigure}{0.35\textwidth}
\begin{center}
\includegraphics[width=0.4\textwidth]{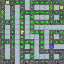}
\includegraphics[width=0.4\textwidth]{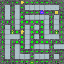}
\end{center}
\caption{Chaser.}
\end{subfigure}\\
\begin{subfigure}{0.35\textwidth}
\begin{center}
\includegraphics[width=0.4\textwidth]{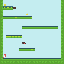}
\includegraphics[width=0.4\textwidth]{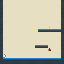}
\end{center}
\caption{Climber.}
\end{subfigure}
\begin{subfigure}{0.35\textwidth}
\begin{center}
\includegraphics[width=0.4\textwidth]{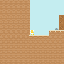}
\includegraphics[width=0.4\textwidth]{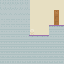}
\end{center}
\caption{Coinrun.}
\end{subfigure}\\
\begin{subfigure}{0.35\textwidth}
\begin{center}
\includegraphics[width=0.4\textwidth]{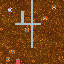}
\includegraphics[width=0.4\textwidth]{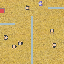}
\end{center}
\caption{Dodgeball.}
\end{subfigure}
\begin{subfigure}{0.35\textwidth}
\begin{center}
\includegraphics[width=0.4\textwidth]{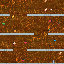}
\includegraphics[width=0.4\textwidth]{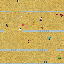}
\end{center}
\caption{Fruitbot.}
\end{subfigure}\\
\begin{subfigure}{0.35\textwidth}
\begin{center}
\includegraphics[width=0.4\textwidth]{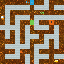}
\includegraphics[width=0.4\textwidth]{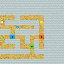}
\end{center}
\caption{Heist.}
\end{subfigure}
\begin{subfigure}{0.35\textwidth}
\begin{center}
\includegraphics[width=0.4\textwidth]{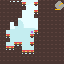}
\includegraphics[width=0.4\textwidth]{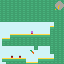}
\end{center}
\caption{Jumper.}
\end{subfigure}\\
\begin{subfigure}{0.35\textwidth}
\begin{center}
\includegraphics[width=0.4\textwidth]{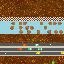}
\includegraphics[width=0.4\textwidth]{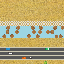}
\end{center}
\caption{Leaper.}
\end{subfigure}
\begin{subfigure}{0.35\textwidth}
\begin{center}
\includegraphics[width=0.4\textwidth]{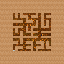}
\includegraphics[width=0.4\textwidth]{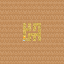}
\end{center}
\caption{Maze.}
\end{subfigure}\\
\begin{subfigure}{0.35\textwidth}
\begin{center}
\includegraphics[width=0.4\textwidth]{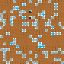}
\includegraphics[width=0.4\textwidth]{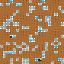}
\end{center}
\caption{Miner.}
\end{subfigure}
\begin{subfigure}{0.35\textwidth}
\begin{center}
\includegraphics[width=0.4\textwidth]{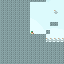}
\includegraphics[width=0.4\textwidth]{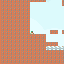}
\end{center}
\caption{Ninja.}
\end{subfigure}\\
\begin{subfigure}{0.35\textwidth}
\begin{center}
\includegraphics[width=0.4\textwidth]{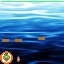}
\includegraphics[width=0.4\textwidth]{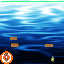}
\end{center}
\caption{Plunder.}
\end{subfigure}
\begin{subfigure}{0.35\textwidth}
\begin{center}
\includegraphics[width=0.4\textwidth]{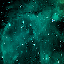}
\includegraphics[width=0.4\textwidth]{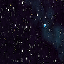}
\end{center}
\caption{Starpilot.}
\end{subfigure}
\caption{Observations from each of the 16 Procgen Environments. Two different domains are shown for each environment.}
\label{fig:procgen-samples}
\end{figure}

Two different domains for each of the OpenAI Procgen environment are depicted in Figure \ref{fig:procgen-samples}. To evaluate the transfer performance of an algorithm, it is trained on one domain and evaluated on the other. Note that Figure \ref{fig:procgen-samples} shows one transfer task for each environment but the results in Figures \ref{fig:cartpole-training}, \ref{fig:procgen-training-easy}, and \ref{fig:procgen-training-hard} are evaluated across multiple trials.

\end{document}